\documentclass[graybox]{svmult}

\usepackage{mathptmx}       
\usepackage{helvet}         
\usepackage{courier}        
\usepackage{type1cm}        
\usepackage{makeidx}         
\usepackage{graphicx}        
\usepackage{multicol}        
\usepackage[bottom]{footmisc}

\usepackage[utf8]{inputenc} 
\usepackage[T1]{fontenc}    
\usepackage{hyperref}       

\usepackage{url}            
\usepackage{booktabs}       
\usepackage{amsfonts}       
\usepackage{nicefrac}       
\usepackage{microtype}      
\usepackage{lipsum}
\usepackage{verbatim}
\usepackage[numbers]{natbib}

\makeindex
\begin{document}

\title*{Evolutionary Computation and AI Safety}
\subtitle{Research Problems Impeding Routine and Safe Real-world Application of Evolution}

\author{
Joel Lehman \\
Uber AI  \\
\href{mailto:joel.lehman@uber.com}{joel.lehman@uber.com} \\
}

\maketitle

\abstract{
Recent developments in artificial intelligence and machine learning have
spurred interest in the growing field of AI safety, which studies how 
to prevent human-harming accidents when deploying AI systems. 
This paper thus explores the intersection
of AI safety with evolutionary computation, to
show how safety issues arise in evolutionary computation and how 
understanding from evolutionary computational and biological evolution
can inform the broader study of AI safety.
}


\section{Introduction}

As the capabilities and pervasiveness of machine learning (ML) and artificial intelligence (AI) 
increasingly affect society, there is
increasing concern about the \emph{safety} of such systems, i.e.\ the potential of accidental harm from implementation errors and unintended consequences in ML algorithms. As a result, there has been increasing interest in the nascent field
of \emph{AI safety} \cite{amodei2016concrete,everitt2018agi,leike2018scalable,yudkowsky2004coherent,christiano2018supervising,irving2018ai}, which seeks to understand and solve the technical challenges
in developing and deploying AI that does what its designer intended it to do. The purpose of this chapter is to explore how the study of AI safety intersects with that of evolutionary computation (EC), to both highlight an exciting and important set of safety problems within EC, and to suggest that evolution and EC have important insights that could benefit the general study of AI safety.

To frame the problem of AI safety, we adopt the framework of \citet{amodei2016concrete}, which defines AI safety as concerned with accidents in ML systems, and defines five problems within three broad categories of issues: (1) specifying the wrong objective function, (2) making safe and efficient use of a true but expensive objective (e.g.\ human feedback), and (3) how to improve or adapt
safely while interacting with the real world. A running example in that paper, which we adopt here, describes a robot with the task of cleaning an office using common tools; we modify the example to assume that the controller for this robot has been evolved, i.e.\ with an EC technique like neuroevolution \cite{lehman2013neuroevolution,yao1999evolving} or genetic programming (GP; \cite{banzhaf1998genetic,koza1992genetic}) in the setting of evolutionary robotics (ER; \cite{nolfi2000evolutionary,lewis1992genetic}). While this running example is posed in the reinforcement learning (RL)
setting of ER, similar issues can arise whenever an EC-trained artifact
interacts with the real world; for example, a credit-scoring system trained with GP symbolic regression (e.g.\ as in \citet{ong2005building}) when deployed might  enact unintended consequences on the real-world borrowers its decisions affect, e.g.\ by basing decisions on ethically- and legally-problematic borrower traits (e.g.\ race).

One motivation for this chapter is to draw attention within EC to a selection of interesting and important  concrete research problems (as introduced by \citet{amodei2016concrete}), in hopes of encouraging progress towards one of EC's aspirations: to provide mature and reliably safe solutions for real-world AI  problems. If EC systems are increasingly trained, refined, and applied in the
real-world, it becomes necessary to deal with real-world complications that are often side-stepped in 
closed-world research benchmarks; grappling with these issues is thus necessary for EC to transition into a reliable approach for safely solving real-world problems.
For example, if evolution is occurring in an environment 
alongside humans (e.g.\ evolving a robot controller that interacts with people in an office setting) much care is needed 
to design an appropriate fitness function that at least does not cause harm in its early incarnations; in contrast,
fitness functions in more traditional closed-world ER simulations often undergo many iterations of free-form debugging, with no real danger or cost (beyond wasted time and computation), where initial attempts often create highly-unexpected outcomes \cite{lehman2018surprising}. To enable reliable real-world deployment of EC, it
may be useful to come up with new automated design procedures, to import tools from AI safety in statistical ML, or to perform new and directed EC research on solving technical safety problems.

A complementary motivation is to highlight AI safety problems for 
which EC techniques might be particularly well-suited to make significant contributions. 
For example, the subfields of
quality diversity (QD; \cite{pugh2016quality,lehman2011evolving}) and open-ended evolution \cite{taylor2016open,standish2003open} 
might provide a natural mechanism to create a diverse set of 
test-scenarios to illuminate rare but important potential failures modes of ML systems (that might
otherwise go unidentified). For example, the fooling images work of \citet{nguyen2015deep} shows how
EC can automatically identify diverse visual patterns that a deep neural network
will confidently misidentify). Overall, while most current AI safety work is
conducted with traditional statistical ML (e.g.\ gradient-based deep learning approaches),
EC might bring new ideas, perspectives, and techniques to bear on such problems.

A final motivation is to consider if and how natural evolution solved problems
similar to those tackled by AI safety researchers. For example, evolution has designed various means of 
collaboration among social animals and between mutalistic species, that in effect minimize negative side-effects to other agents (an important topic in AI safety). 
Additionally, evolution has uncovered ways to explore more safely both across an evolutionary timescale (i.e.\ through the evolution of
evolvability \cite{kirschner1998evolvability,wagner1996perspective}, whereby evolution favors improved variation) and an individual organism's lifetime (i.e.\ through the complementary instincts of curiosity and fear \cite{buss2015evolutionary}). The hope is that biological inspiration might point the way towards potential solutions to these kinds of safety problems in EC or in ML at large.

The conclusion is that AI safety is likely to be a growing field of interest in
coming years that offers a range of interesting technical challenges, 
and that EC may both have important insights to offer and benefits to gain from research in that community.


\section{Background}

The next sections describe the field of AI safety, and how EC is applied
in the real world, which helps to understand safety concerns from an EC perspective.

\subsection{AI Safety}

The field of \emph{AI safety} \cite{amodei2016concrete,everitt2018agi} seeks to pose and solve  technical challenges involved in developing AI that in practice does what its designer intends it to do. The hope is to help foresee and avoid harmful accidents that might result from good-intentioned AI gone astray, for example, through misspecified fitness functions or differences between the training and testing environments.
While the name ``AI safety'' naturally evokes ideas of direct physical safety (e.g.\ how to make sure there are sufficient guard-rails that prevent a robotic arm from accidentally hitting a human), the problems studied in AI safety also encompass more abstract and broad concerns. Such concerns include immediate and short-term ones, like how a mobile robot driven by RL can continually improve its policy by exploring, without taking any catastrophic actions (such as those that cause harm to itself, to the environment, or humans); they also include more speculative concerns about the future (e.g.\ how to make sure
an AI that surpassed human intelligence would still be controllable and aligned with our interests).

One central challenge in AI safety, relevant both to short and long-term concerns, is known as the \emph{value alignment} problem: How to align what a computational agent values with what we value. This problem might appear at first simple, because as designers of agents we have complete control over their incentives. However, such alignment remains an unsolved technical
challenge. Currently we do not know how in practice to algorithmically specify (or learn from data) the complexity of what humans care about, e.g.\ our
moral intuitions, common-sense knowledge, and cultural norms, all of which can potentially come to bear upon what we intend for a computational agent to do.
In other words, EC as of yet lacks a  procedure to specify a correct and complete fitness function that encompasses all the background context that could be important
for a system that interacts appropriately with humans and society. 

More concretely, even for an AI system that interacts with the real world in very limited ways, it is
still often a challenge to design a fitness function that truly measures or incentivizes correct behavior \cite{lehman2018surprising}.
Indeed, the typical paradigm in AI remains to specify a fixed and
relatively simple objective function (e.g.\ a fitness function in EC) that is then optimized through search; however, as practitioners in EC are well-aware, an intuitive fitness function can often be optimized in unexpected ways \cite{lehman2018surprising}. While there exist candidate approaches to value alignment \cite{irving2018ai,leike2018scalable,yudkowsky2004coherent}, the problem at core currently remains unsolved.

Interestingly, even if incentives are aligned, i.e.\ the learning system is provided
with the correct objective function, how to successfully (and
safely) optimize that objective function is still a difficult and unsolved problem
in its own right. For example, an RL agent that is given the correct objective
to optimize can still make mistakes \emph{while it is being optimized} (e.g.\ it can make harmful mistakes
while exploring how to improve its policy); or, the objective might be challenging to optimize
(e.g.\ it might instantiate a fitness landscape with many local optima), and the
locally-optimal policies found by search in practice might not be value-aligned.

One useful framework for categorizing technical challenges in AI safety 
comes from \citet{amodei2016concrete}, which divides safety problems into five categories:  avoiding negative side effects,
reward hacking, scalable oversight, safe exploration, and robustness to distributional shift (see table \ref{tab:tcais} for short descriptions of each).
We adopt this framework in this paper for relating AI safety problems to EC and evolution, and later in this paper describe each of these
problems in detail and how they emerge in EC.

\begin{table}[]
\begin{tabular}{|p{4.2cm}|p{7.3cm}|}
\hline
            Avoiding Negative Side Effects &  Negative side effects result from a fitness function that correctly specifies how to \emph{narrowly} achieve a goal, but does not penalize possible harms to the environment or other agents.           \\ \hline
            Reward Hacking & Reward hacking is when a fitness function fails to well-specify how to achieve a goal; evolution can therefore maximize fitness in an unexpected and undesirable way.            \\ \hline
            Scalable Oversight &  Scalable oversight requires effectively and efficiently balancing the use of cheap proxy fitness functions (e.g.\ a simple heuristic) with expensive but more accurate fitness evaluations (e.g.\ human assessment).  \\ \hline
            Safe Exploration & Safe exploration studies how evolution can learn effective behavior while minimizing catastrophic actions taking during learning. \\ \hline
            Robustness to Distributional Shift & Robustness to distributional shift requires real-world applications of evolution to safely deal with situations not seen during training. \\ \hline
\end{tabular}
\caption{The table describes five categories of technical challenges in AI safety, as identified by \citet{amodei2016concrete}.}
\label{tab:tcais}
\end{table}

One general consideration for AI safety is that it is most relevant when considering applying AI algorithms to real-world
situations, where human well-being, broadly speaking (e.g.\ including not only physical safety, but also social harm from biased high-stakes decisions \cite{whittaker2018ai} or offense from insensitive classifications \cite{usatoday}), might be at stake. Thus the next section reviews common paradigms for applying EC to the real world.

\subsection{EC and the Real World}

\begin{figure}
    \centering
    \includegraphics[height=2.4in]{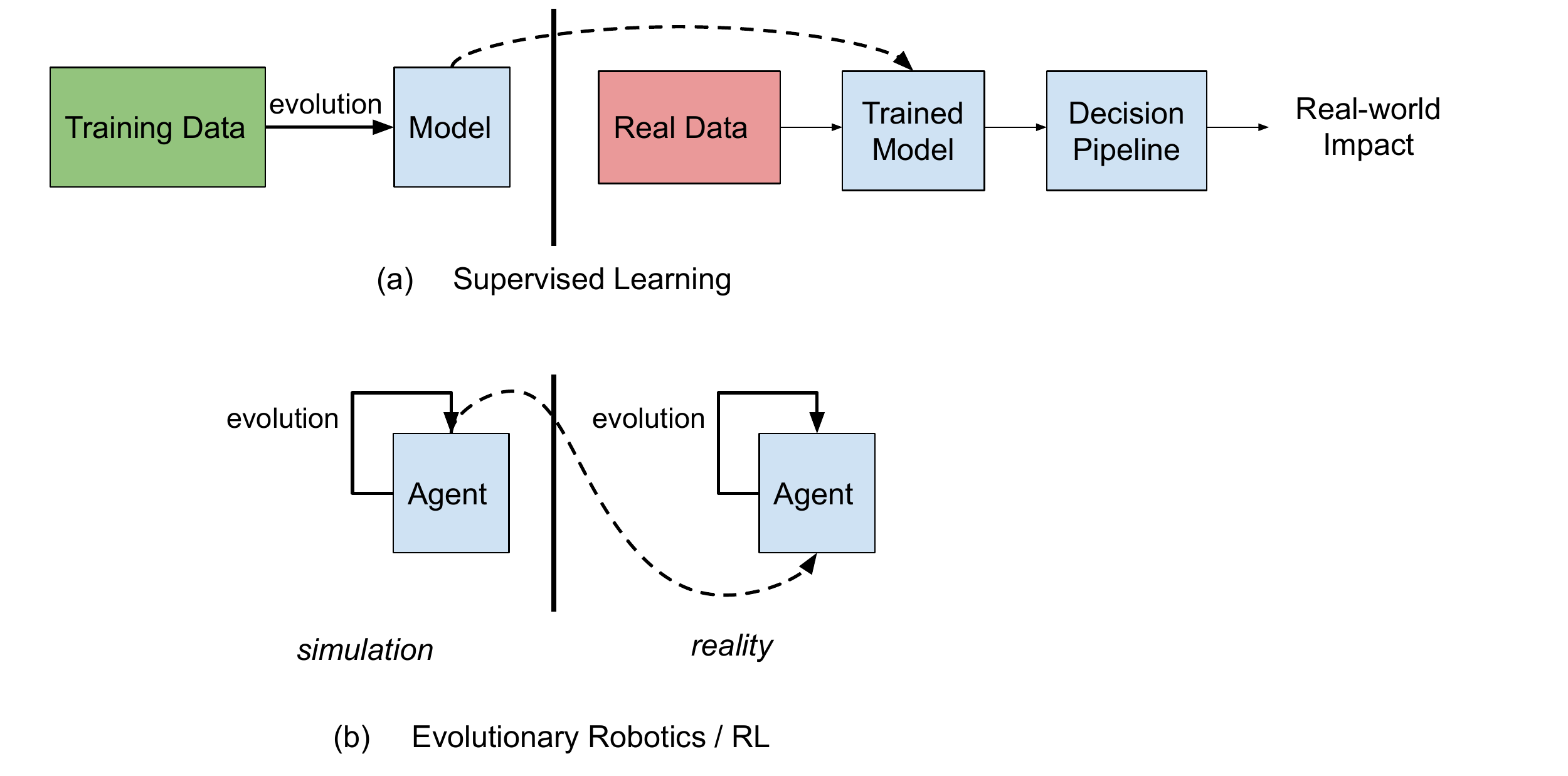}
    \caption{Common paradigms for EC to impact the real world. In (a) supervised learning, a model is evolved by fitting training data, and then is deployed into a larger decision pipeline that involves real data; the decisions resulting from the pipeline (that are influenced by the trained model) translate into real world impact. In (b) evolutionary robotics or RL, an agent is often first trained in simulation, and then transferred across the reality gap, where it can potentially be further evolved. Alternatively, in embodied evolution, an agent is trained from the onset in the real world.}
    \label{fig:ecrw}
\end{figure}

There are many different motivations for studying EC. While one entirely legitimate 
such motivation is to understand the creative potential of
algorithms inspired by biological evolution for its own sake, researchers in EC often explicitly 
aim towards real-world applications of their ideas, or at least paint a viable path towards
how their ideas might be translated into beneficial real-world impact. Below we describe how such translation
often happens in both supervised and reinforcement learning problems (see figure \ref{fig:ecrw} for a high-level summary).

\subsubsection{Supervised Learning}

When EC is applied to supervised learning, i.e.\ training where the task is
to predict or classify over a labeled training set, it is important to recall that
supervised training performance 
is rarely an end in itself. While improved accuracy on a benchmark is often critical
 for publishing a paper about symbolic regression \cite{koza1994genetic} or neural classification \cite{rocha2007evolution}, such accuracy is only practically important insofar as it feeds into the downstream task the model is \emph{applied to}. For example, a classification model of credit-worthiness
might be applied to decide whether a loan should be granted or not. While improved accuracy will likely
contribute to such use cases, it will not take into account the nuances of the domain, like the differing impact of different kinds of mistakes \cite{usatoday}.

Thus, while applications of supervised learning might
at first not seem relevant to AI safety, the objective that a supervised-learning EC model is
trained towards (e.g.\ classification accuracy) nearly always serves only as a \emph{proxy} for the ``true'' downstream objective in the real world
(e.g.\ efficient loan allocations that abide by legal and moral norms). Notably, limitations
of such proxies are well-known; for example, the fairness, accountability, and transparency
community within ML has highlighted how maximizing training accuracy can result in models that
base decisions on societally-unacceptable criteria \cite{zafar2017fairness,whittaker2018ai}. This pervasive type of gap between the proxy and the true objective can
be seen as a manifestation of the general value alignment problem, and techniques for minimizing such a gap
highlight how AI safety research can be relevant to EC-based supervised learning.

\subsubsection{Reinforcement Learning}

When EC is instead applied to robotics or RL, evolution most often first occurs within
simulated environments. The idea is that policies trained in simulation can subsequently be transferred to 
reality \cite{jakobi1995noise,koos2013transferability}, and potentially further evolved in the real world. The reasons for training
in simulation include that real-world evaluations can be slow, tedious, and expensive, often
risking damage to hardware (like a robot) and to the broader environment (like humans coexisting with the robot). Simulation enables more convenient large-scale 
experimentation (given sufficient computation), although both how to design accurate simulations for complicated domains and how to successfully transfer policies from simulation
to the real world remain challenging areas of research \cite{pollack2000evolutionary,koos2013transferability,jakobi1995noise}. 

Safety concerns in this paradigm can emerge from simulations that do not reveal safety-critical edge-cases later encountered when models are deployed in the real-world, or from changing circumstances in reality (i.e.\ distributional shift) that are not captured for in simulation.
Another paradigm in EC is embodied evolution \cite{watson2002embodied}, wherein evolution is conducted in the real world, to circumvent the challenges of
building accurate simulators and crossing the reality gap. In this setting, to the extent that evolved policies interact with humans or can damage their
robotic body or their environment, there may be the need for potentially expensive supervision (an AI safety issue discussed in more detail later). In
general, because there is never the protective buffer of simulation between a policy and the real-world, safety considerations in embodied
evolution may be more challenging than in other settings.

The conclusion is that as EC strives and achieves greater real-world impact, there will likely be a corresponding increased risk (albeit still potentially minor in many domains) of unintentional harm,
independent of the specific paradigm by which EC models are trained and deployed.

\section{EC and Concrete AI Safety Problems}

This section explores more concretely how ideas from EC intersect with those from AI safety. We adopt the
framework of \citet{amodei2016concrete}, which identifies five classes of concrete problems that can cause AI accidents: avoiding negative side effects,
reward hacking, scalable oversight, safe exploration, and robustness to distributional shift (described from a high level in table \ref{tab:tcais}). 
For each of these five problems
we introduce the problem, describe how it can arise in EC, how it relates to various research areas in EC, and suggest directions for potential solutions to such problems. 
Note that our main aim here is to frame AI safety for EC researchers and practitioners, and as a result, we will not comprehensively survey the broader study of AI safety within ML; for more comprehensive surveys, see \citet{amodei2016concrete} or \citet{everitt2018agi}.


\subsection{Avoiding Negative Side Effects}

The problem of negative side effects is that a well-specified fitness function must not only reward achieving a desired goal narrowly, but should also penalize possible negative consequences on the broader environment. That is, a fitness function is often under-specified in practice, even if the conditions of achieving the desired goal are correctly described. The reason is that there are many ways to short-sightedly accomplish a goal that humans would nonetheless find unacceptable. For example, borrowing from \citet{amodei2016concrete}, a robot might knock over an expensive vase en route to its destination; even if the robot arrives successfully at its destination (its goal), the damage to the vase is an unacceptable negative byproduct of the robot pursuing its goal. 

If the fitness function does not penalize for breaking the vase, the resulting negative side effects could be viewed as a failure of the researcher to express the correct fitness function. However, while one could
attempt to anticipate and hard code into the fitness function every negative contingency, such exhaustive anticipation is often unrealistic, and at best tedious. Ideally, there would be a way to
automatically (or with minimal supervision) augment a goal-directed fitness function to penalize such undesired impacts.
The challenge in designing such an automated method relates to the value alignment
problem in AI safety, in that there is much background context (e.g.\ about what objects in the environment are fragile or important) that a human brings to their understanding of what an acceptable solution is; such context is difficult to effectively and exhaustively translate into a fitness function (although some projects do aim to distill such background knowledge \citep{lenat1995cyc}).


Interestingly, most EC and ER environments are constructed such that there is little potential for 
negative side effects; the reason is that richer environments are more challenging to model, more computationally demanding to simulate, and
such complications are often orthogonal to the
research questions under study. In practice, simulated environments in ER are nearly always 
closed-world and spartan, containing only elements directly relevant to the task at hand. 
For example, a common variety of ER task involves simulated wheeled robots navigating through
an enclosed environment containing only walls and artifacts directly related to the task (e.g.\ 
a light switch that can be triggered, or tokens that can be collected). Negative side-effects
are often impossible by definition: The robot can not damage itself or anything of importance in
its environment. 

In ER and EC experiments that involve the real world, or interacting with humans, there is
more potential for negative side-effects, although experimenters nearly always apriori
minimize that possibility by design. For example, when transferring policies evolved in simulation 
to the real world, the real world environment is often engineered to mimic the spartan simulated one,
and often such transfers are one-off experiments (i.e.\ the robot will not then be operating
in an ongoing way) under intensive supervision.
However, despite the minimization by design of negative side effects, 
the conclusion is that as (or if) EC and ER progresses, we likely will want or need 
evolved agents to be deployed in
complex open-world or human-coinhabited environments; in such situations, the problem of 
negative side effects can no longer be avoided. Thus, 
when aiming toward the real world, simulated environments may need to
be augmented to include the potential for negative side-effects (and for learning
to avoid them), or automated techniques for mitigating side-effects from
real-world deployment may need to be developed.


So far, the problem of negative side-effects appears to be an
under-studied aspect of how to scale EC, one that may provide exciting future research
directions. One possible paradigm for minimizing negative-side effects is to train
EC agents through interactive evolutionary computation (IEC; \cite{takagi2001interactive}), i.e.\ to
involve humans directly in the breeding process. Due to the problem of user fatigue in IEC \cite{takagi2001interactive}, i.e.\ that
the task of breeding can become monotonous and exhausting,
it is difficult to scale IEC, which necessitates learning surrogate models \cite{jin2011surrogate} or applying distributed IEC \cite{secretan2008picbreeder}, i.e.\ systems that involve many humans breeding in potentially uncoordinated ways. 
Overall, the interaction of IEC with embodied evolution or
ER in general (as in \citet{woolley2014novel}) could benefit from greater study
from a safety perspective. Current research directions in ML that address negative side effects
include penalizing for changes to the environment \cite{armstrong2017low}, or
algorithms that \emph{satisfice} instead of optimize unboundingly \cite{taylor2016quantilizers} (motivated by the idea
that side-effects may often result from extreme optimization). Both such approaches could potentially
be adapted for EC.

\subsection{Reward Hacking}

The problem of reward hacking, like that of negative side-effects, is caused by an
incompletely- or incorrectly-specified fitness function. While negative side-effects
are collateral damage incurred while successfully achieving the desired objective, reward hacking is
when optimization uncovers unexpected ways to maximize the fitness function \emph{without}
 achieving the desired objective. For example, if the true objective
of a cleaning robot is to clean the office, but its fitness function rewards
 for each individual mess the robot cleans, the robot may discover that it maximizes fitness by
 creating new messes that it can subsequently clean \cite{amodei2016concrete}.


The phenomenon of reward hacking is familiar to most EC practitioners; nearly all of us
have encountered situations where an intuitive fitness function is maximized by 
counter-intuitive (and undesirable) behavior. Indeed, that so many illuminating (and funny) anecdotes of reward
hacking existed in the EC community was one main inspiration behind the crowd-sourced documentation effort of \citet{lehman2018surprising}, which describes many reward-hacking examples. A representative example is found in
 Karl Sims' seminal virtual creatures work \cite{sims1994evolving}. In early attempts to evolve locomotion gaits by rewarding forward motion, the result was not locomotion, but morphological evolution towards tall rigid bodies that could exploit their potential energy by falling or somersaulting forward.

Beyond EC, the challenge of constructing incentives for agents (like fitness functions) that cannot be
undermined is well known in other fields. For example, in economics, Goodhart's law \cite{goodhart1984problems} states 
that ``when a measure becomes a target, it ceases to be a good measure.'' Similar understanding goes by the name of the principle-agent problem
in economics and political science \cite{ross1973economic}, and similar challenges exist in designing contracts in law \cite{hadfield2018incomplete}. Further, there
are many historical examples of perverse incentives, where an incentive to solve one problem instead exacerbates it; for example, a French colonial program in Hanoi paid citizens for turning in rat tails, in hopes of exterminating rats, but it instead led to \emph{farming} rats \cite{vann2003rats}. 
This consilience of evidence suggests that designing incentives is generally difficult, and that
humans are habitually overconfident about their ability to skillfully do so, often failing to
anticipate subtle loopholes instantiated by intuitive reward structures. 
In this way, reward hacking in EC and ML is one manifestation of a broader problem.

In practice, reward hacking in EC is often solved through iteration. First, an intuitive fitness
function leads to surprising and undesirable outcomes, that are understandable only in hindsight. The 
experimenter then attempts to modify the fitness function to patch the problem, which
potentially may lead to a different kind of exploit that must also be patched. 
Interestingly, because these failed incentives can be viewed as failures of the
experimenter, and happen within the loop of scientific experimentation that precedes
a polished experimental setup, 
they are often not reported scientifically \cite{lehman2018surprising}; 
as a result, the prevalence and 
importance of reward hacking in EC may be under-appreciated and understudied. 

While frustrating, when evolution occurs in simulation 
such reward hacking may not cause harm much beyond wasted experimenter
effort and time. However, the ability for EC practitioners to quickly and
safely explore new tasks, especially in settings such as embodied evolution or 
reality-gap crossing, is undercut by the expertise and trial-and-error needed 
to construct reliable fitness functions.

As in negative side-effects, IEC is one avenue for helping to overcome
reward hacking, by involving human judgment to assess quality during
evolution
rather than by crafting fixed heuristics. Beyond directed human breeding, humans may also supply other (potentially richer) forms of guidance to further constrain or replace traditional
fitness functions, like demonstrations of acceptable behavior or heuristic advice, as in \citet{karpov2011human}. 

Such EC research directions can be seen as connected to similar potential solutions
in traditional ML, such as imitation learning \cite{argall2009survey}, 
wherein an agent learns how to imitate expert demonstrations of behavior; cooperative inverse reinforcement learning \cite{hadfield2018incomplete}, where a reinforcement learning
agent cooperates with a human to discover and optimize the human's preferences; or reward modeling \cite{leike2018scalable}, wherein a machine learning model is trained to
predict human preferences (similar to surrogate models used in IEC \cite{jin2011surrogate}. 
Exploring if and how such ML methods could apply to EC (e.g.\ evolutionary imitation learning,
or applying deep learning models to learn models of human preferences to drive evolution) may
be a productive area of future research.


\subsection{Scalable Oversight}

The problem of scalable oversight is that in EC and learning systems in general, there
is often expensive-to-gather information that accurately reflects how acceptable a
solution is, but such guidance is too expensive to be applied as the primary
driver of search. For example, a very accurate measure of fitness for 
a cleaning robot might require expensive manual testing of how dirty
a carpet is before and after the robot is deployed within a room. Other
proxy measures may be more cheaply available, such as a human giving a quick glance to a room, or
by the robot measuring how much dirt it is picking up. However, such proxies
might exacerbate problems such as negative side effects or reward hacking \cite{amodei2016concrete}. For example, a robot maximizing dirt picked up might knock over a plant to gain access to more dirt, or a robot maximizing human approval after a quick glance might hide messes under a rug.
The issue is how to efficiently and effectively apply combinations of cheap
proxy signals with occasional expensive feedback, to produce a practical (and well-behaved)
learning system.

One way the issue of scalable oversight emerges in EC is through
the practical construction of real-world fitness functions (e.g.\ fitness functions for
fine-tuning policies in reality that were first learned in simulation, or fitness functions applied in embodied evolution). In other
words, when applying evolution in a real-world situation, what sensors are available on a robot, what a human can easily evaluate, or how the environment can be augmented with automated sensors to evaluate aspects of behavior (e.g.\ motion capture equipment or ceiling-mounted cameras) will affect what
fitness functions are possible to automate, and the overall cost-effectiveness of executing different experiments. 

However, scalable oversight, like other AI safety issues, is often eliminated by
design from simulated EC domains. Experiments in which cheap proxy fitness evaluations are not possible
or in which they fail (due to reward hacking or negative side-effects) are unlikely to be pursued or published. However, if progress could be made on
enabling more scalable oversight, it might extend the range of what kinds of embodied evolution
or real-world fine-tuning could be performed. In this way, scalable oversight is an interesting avenue of research not
only for safety reasons, but because it may help expand the complexity of domains for which real-world EC can
be applied.

The area of EC research most similar to scalable oversight is that of surrogate-assisted EC \cite{jin2011surrogate}, wherein expensive-to-calculate fitness functions are approximated
with a learned model; of particular interest (for their potential efficiency) 
are surrogate models that intelligently
choose which points in the search space to subject to expensive ground-truth fitness queries. 
For example, \citet{gaier2018data} applies Bayesian optimization to enable a data-efficient quality diversity \cite{pugh2016quality} algorithm.  Another related area of EC study are methods that estimate which
genomes are likely to successfully cross the reality gap \cite{koos2013transferability}; the
reason is that there is an analogy between simulations (and their relation to reality) and
proxy fitness measures (and their relation to ground-truth fitness).

From a ML perspective, \citet{amodei2016concrete} propose that a semi-supervised formulation of reinforcement learning may be a productive paradigm for tackling scalable oversight. The idea is
that an agent only receives reward information on a small subset of its experience (as opposed to
the traditional RL setting where reward is observed for each action taken in the environment). In
particular, the agent must learn \emph{when} to request expensive reward information, and is
incentivized to learn cheap proxy measures that correlate with the expensive reward. Because EC
uses fitness functions that operate over an individual's entire evaluation, rather
than the per-timestep rewards of traditional RL, it may not be easy to translate
such a paradigm to EC (although it could be an interesting direction for research). 
One potential way of framing semi-supervised RL for 
evolutionary RL is to learn a semi-supervised reward predictor (with ML) that could
assign fitness to individuals by observing their sensory-motor stream.




\subsection{Safe Exploration}

The problem of safe exploration is how evolution (or individuals capable of life-time learning) can explore new solutions without ever (or only very rarely) taking catastrophic actions, i.e.\ ones that harm valuable aspects of the environment, including humans or expensive equipment such as robots. 
Note that safe exploration remains a problem even if objectives are correctly specified: Even if a fitness
function correctly identifies all unacceptable negative side-effects, and a \emph{properly-trained} agent would thus avoid such effects, \emph{during learning} an agent might still undertake catastrophic actions. For example, 
the cleaning robot may suffer a fitness penalty for breaking a vase, but it still needs to experience that penalty during training to learn to avoid breaking it. A related problem is that given a robotic controller that behaves safely, there is no guarantee that an arbitrary mutation of it will also
be safe. The danger of exploration is a deep philosophical problem, in that the very act of exploration seems inherently to be about stepping into the unknown. However, humans can often
successfully explore new possibilities and emerge relatively unscathed (sometimes using mental models to predict whether a new strategy would be catastrophic before trying it, somewhat similarly to model-based RL \citep{sutton1990integrated}), suggesting that practical solutions may be possible.


There are two main ways that real-world 
accidents from safe exploration can emerge in EC. First, take the case of learning a plastic policy (e.g.\ a policy that learns from experience \emph{during its lifetime} \cite{soltoggio2008evolutionary,soltoggio2018born}). For example, a robot might be trained to explore any environment it is embedded within, in search of a particular goal. In effect, such an agent must
learn \emph{how to explore}, and if the deployment plan involves the real world (through embodied evolution, or crossing the
reality gap), then there are risks from unsafe exploration. For example, in a new environment, a learned exploratory strategy might
lead the robot to damage itself. Second, there is the case where a learned (non-plastic) policy is either trained
in the real world (embodied evolution), or is fine-tuned in the real world after being trained in simulation. In this
case, exploring the space of policies (through mutations of existing policies) may result in unsafe policies. For example,
in some robotics domains solutions are known to be fragile, i.e.\ that most mutations result in
degenerate (possibly damaging) behavior \cite{lehman2011improving,lehman2018more}. For concreteness, a robot trained to walk successfully
in simulation may lose some performance when transferred across the reality gap, and there is no guarantee that 
perturbations of the transferred policy (explored in hopes they will improve the walking policy) will not cause the robot to fall and harm itself.

Overall, it may be impossible to solve the issue of safe exploration without involving
some form of human oversight. The reason is that learning what is unsafe 
seemingly requires either: (1) an accurate model of the world that includes robust identification of
catastrophes, (2) labelled data of all possible causes of unsafe scenarios in a domain, or (3) active experience
in the domain with feedback from an overseer that prevents unsafe actions from being taken. 
All three require either extensive domain knowledge, e.g.\ (1) or (2), or direct human intervention (3).
In this way, the problem
of safe exploration may be intrinsically tied (like some of the other problems) to that of scalable oversight: Given
that potentially expensive human feedback is needed, how can it be gathered and exploited in an efficient way to 
enable reliable real-world exploration? 

Interestingly, 
like other problems mentioned here, often the issue of safe exploration in EC currently arises \emph{outside} the formal
scientific process: Domains are constructed that intrinsically minimize risk (e.g.\ through spartan closed-world design), and guard-rails to minimize damage to real-world robots and their environment are
engineered on a robot-by-robot or domain-by-domain basis by experimenters; failure modes (e.g. robot damage) encountered in such experiments are unlikely to be deemed of enough scientific import to be published. Thus, one contribution to studying safe
exploration in EC would be to introduce a variant of common ER benchmarks that simulate the idea of safe
embodied evolution; for example, a maze navigation task could include deep holes that would endanger a robot,
or fragile and valuable aspects of the environment. 

Another possible avenue of research for contributing to safe evolutionary exploration is to
improve the robustness and evolvability of genomes. For example, some EC methods find parts of the search space that are more robust to mutation \cite{lehman2018more}, or adapt variation operators to increase robustness or evolvability \cite{lehman2011improving,wierstra2008natural}, or attempt to enforce small changes to an evolved policy \cite{lehman2018safe}. While not initially motivated by safe exploration, it may be possible to adapt such techniques towards that end. The idea is that with well-tuned variation, parent policies that are safe may be more likely to produce safe children policies, under the assumption that larger policy changes are more likely to be degenerate.

EC could also attempt to solve existing safe exploration
benchmarks from the RL community, e.g.\ the safe exploration grid-world of \citet{leike2017ai} or
domains explored by \citet{moldovan2012safe}.
Potential safe exploration techniques could also be imported or adapted from studies of safe exploration in RL \cite{garcia2015comprehensive}. Promising such techniques include the
approach of \citet{saunders2018trial}, wherein human oversight is used to train a supervised learning model
that blocks unsafe actions, or \citet{lipton2016combating}, wherein catastrophic actions are
explicitly stored and rehearsed to endow a RL agent with an intrinsic sense
of fear. Similar models could be trained to block unsafe actions for ER or
in embodied evolution.

\subsection{Robustness to Distributional Drift}

The problem of robustness to distributional shift is that when AI systems are deployed, they may  encounter situations that deviate from
the ones it was trained upon. In such situations, a naively trained agent may demonstrate
arbitrarily inappropriate behavior, because extrapolating to novel circumstances is challenging.
Accidents can thus result in this paradigm if an agent's
policy results in ill-suited actions when encountering new situations.


In some EC communities, such as ER, experiments may not always explore how well a learned behavior
generalizes to situations other than the exact ones experienced in training; i.e.\ 
in the language of statistical ML, the training set doubles as the testing set. 
As a result, there may be little understanding of how a policy would generalize, and
how pathological a robot's behavior would be if it encountered a novel situation. 
Note that interestingly, the issue of poor generalization 
is a topic of recent interest in deep RL 
as well \cite{zhang2018study,justesen2018procedural,cobbe2018quantifying}.

While this paradigm may not be intrinsically problematic, i.e.\ if the research
question does not involve generalization or real-world deployment, 
graceful degradation of out-of-training-distribution performance becomes critical as policies are
deployed in the real-world (especially open-world scenarios where it is well-understood that all possible situations cannot be anticipated, and that circumstances will likely shift over time). 

Several EC communities study partial solutions to 
this problem. For example, one subfield of EC studies
dynamic fitness landscapes \cite{branke2003designing,richter2009detecting},
wherein evolution continues as circumstances change, which could continually align the policy to the current
distribution of scenarios. Further, such fluid adaptation may favor (or be enabled by mechanisms that encourage) more \emph{evolvable} representations, i.e.\ representations offering diverse and adaptive variation, another important and related field of EC study  \cite{wagner1996perspective,kashtan2007varying}. Complementarily, others in EC study meta-learning \cite{soltoggio2008evolutionary}, i.e.\ evolutionary approaches to learning \emph{how to learn}, which may enable a policy to quickly learn online from its own mistakes. 

While these research communities provide
important insights for tackling distributional shift, new benchmark tasks may
be needed to ground out the risks from real-world distributional shift and to determine which (or which
combinations) of these techniques would help ameliorate such risks in practice. 
For example, an ER domain could be introduced in which environments are produced
through procedural content generation (PCG; \cite{shaker2016procedural}), but where the
distribution of PCG parameters changes over evolutionary time; different approaches
could be compared by how many catastrophic failures are encountered across
evolutionary time.

Solutions could also take inspiration from the study of distributional shift
within ML. For example, the insight in Inverse Reward
Design \cite{hadfield2017inverse} is that the
fitness function
encountered during training should only be trusted insofar as
it reflects situations that occur during training (i.e.\ the human
designer of the fitness function designed it explicitly to solve
such training situations). An agent
should thus have uncertainty over what such a fitness function implies for 
for situations that never appear in training environments. It may be possible to export such an 
insight to an evolutionary context, perhaps by querying a human for guidance or forcing
a known safe policy to take over when anomalous circumstances
are encountered (e.g.\ as studied by the fields of novelty/anomoly detection \cite{markou2003novelty,chandola2009anomaly} or
uncertainty-aware RL \cite{kahn2017uncertainty,eysenbach2017leave}). 

\section{Discussion}

One interesting question is if EC has unique contributions to make to the 
general study of AI safety. A potential benefit of evolution relative to
traditional ML is its divergent creative potential -- evolution seems
well-suited to discovering a great diversity of well-adapted artifacts. Subfields of EC that
study artificial life \cite{langton1997artificial}, open-endedness \cite{standish2003open}, and quality diversity \cite{pugh2016quality} focus on
this facet of evolution, which may be of use for helping in particular
with the problem of robustness to distributional shift. That is, evolution
could be driven to discover a wide range of new training situations to
discover latent flaws in learned policies or models, to augment a limited
training set that might not cover the diversity of situations that could later
be encountered. For example, the
work of \citet{nguyen2015deep} applies a QD algorithm to find, in a
single evolutionary run, a set of diverse images that reliably fool
a deep neural network vision model; following work has shown that
these kinds of adversarial images can provide safety hazards for
real-world use cases of such vision models \cite{eykholt2017robust,kurakin2016adversarial}.
Similar QD approaches might also be used to evolve scenarios to stress-test robotic policies.
Work in this spirit includes \citet{goldsby2010automatically}, wherein
novelty search and GP are used to probe latent behavior of a robotic navigation
system and an automobile door locking control system. Similarly, the environments evolved by
open-ended systems like POET \cite{wang2019paired} could be adapted as a testing suite 
for fixed policies.

A related question is to consider what lessons biological
evolution has for AI safety. Many problems faced by AI safety have been
solved, at least in some abstract sense, by biology. For example, the 
problem of negative side effects in AI safety is related to the evolution of cooperation and sociality in biology, in
 that cooperation often entails considering other agents and their goals in addition
to one's own goal (whether through behavioral convention, as in bees, or deliberative thought, as in humans). From this perspective,
the negative side effects of a robot pursuing its own limited agenda
result from not understanding or taking into account the broader preferences of outside agents
(e.g.\ that a vase is a valuable artifact and should not be broken while cleaning a room). Humans have evolved
moral instincts, the ability to empathize with others, and verbal and written language, all of which enables us to
understand the gestalt of a task another human might ask us to perform, thereby helping us avoid reward hacking and
negative side effects. Similarly, the robustness of our
genetic architecture to random mutations and the natural instincts of
curiosity and fear are nature's hard-won solution to the problem of safe exploration on a genetic and individual level, respectively. 
In the same way that evolution (and EC)
have a privileged position in the study and understanding of human-level AI (because evolution is the only algorithm to so-far produce human-level intelligence), evolution and EC may also have a privileged position in understanding the AI safety challenges that
biology has in some sense solved.

An important question for future study 
is if methods in EC manifest different kinds of AI safety concerns than those
considered within traditional ML, e.g.\ due to their lack of formal gradient-following or because some EAs produce AI as the result of a divergent creative process (as opposed to optimizing an explicit
objective function as common in most ML).
Because this question is yet unanswered, it is unclear whether the long-term safety agendas currently popular in ML \cite{leike2018scalable,christiano2018supervising,irving2018ai}
are applicable to AI produced by paradigms such as evolutionary artificial life or open-ended evolution, which in their grandest aspirations (just as in traditional
ML or AI) include producing agents with human-level intelligence \cite{stanley2017open}. If current safety agendas do not apply to ambitious forms of EC, then formulating new agendas that targeting them may be a valuable pursuit.

A final discussion topic is to draw together some of the recurring themes from considering each AI safety problem separately, in hopes of
highlighting promising research questions and paradigms.  One theme is the
potential need for modifications of EC benchmarks to include safety considerations or the adoption of existing AI safety benchmarks within EC.
Benchmarks, for better or worse, help draw researcher attention,
and can render seemingly nebulous problems more concrete. Because existing EC domains and benchmarks minimize safety concerns by design (because
researchers most often are pursuing research questions orthogonal to safety), new benchmarks may help to catalyze safety research, especially
if they are variants of domains familiar to EC researchers.
For example, EC techniques could be applied to the AI safety grid-worlds of \citet{leike2017ai}. Alternatively, existing ER domains (such as maze
navigation or ball-gathering) could be augmented with catastrophic actions (for investigating safe exploration), or could include held-out
test environments that could test for robustness to distributional shift.
Another overarching theme is the potential for some form of IEC to help in the
solution to nearly all of the reviewed problems; this is not surprising, because many AI safety problems emerge precisely because
human insight is relegated to constructing a fixed setup (i.e.\ in EC the genetic encoding and the fitness function), and IEC is a
framework for allowing human choice to intervene during evolution. Safety considerations may drive more efficient ways to perform
IEC (through improved surrogate models), as well as the construction of new forms of IEC. For example, IEC most often helps steer
what individuals reproduce, but IEC solutions to problems such as safe exploration may require humans to interact more directly
with policies \emph{as they execute}, i.e.\ to intervene to prevent unsafe actions. One source of inspiration may be systems
such as the neuroevolution-based game NERO \cite{stanley2005real}, in which a human experimenter can interact in real time to dynamically change the
environment, parameters of the fitness function, and even embody a virtual agent to probe learned agent behaviors.

\section{Conclusion}

AI safety is an important research topic for enabling EC to reach one of its aspirations, which is to maximize its beneficial real-world impact. At first glance,
such research might seem uninteresting, because it can evoke sentiments of domain-specific engineering, rather than the pursuit of grand scientific questions; however,
AI safety enfolds interesting and philosophically deep unsolved technical challenges, including how to avoid catastrophe while learning about the world, and how to create
fitness functions that incentivize agents that abide by the spirit rather than the letter of the law. As ML and AI grow in import, we can expect funding and interest in AI safety to similarly grow,
and the hope of this paper is to advocate for EC researchers to both contribute and take note of advances in this developing field.

\bibliographystyle{unsrtnat}  
\bibliography{lehman_references}  

\end{document}